\documentclass[letterpaper, 10 pt, conference]{ieeeconf}  

\IEEEoverridecommandlockouts                              

\let\labelindent\relax
\usepackage{enumitem}
\usepackage{amssymb}
\usepackage{graphicx} 
\usepackage{cite}
\usepackage{amsmath}
\usepackage{booktabs}
\usepackage{multirow}
\usepackage{xcolor}
\usepackage{pifont}

\usepackage{xspace}

\newcommand{\eg}{\textit{\textrm{e.g.}}}
\newcommand{\etc}{\textrm{etc.}}

\newcommand{\cmark}{\ding{51}}%
\newcommand{\xmark}{\ding{55}}%

\def\fr{InFeR}
\def\frinet{\fr{}-INet}
\def\frdec{\fr{}-DEC}
\def\frgnm{\fr{}-GNM}
\def\undec{\fr{}-DEC-Blind}
\def\ood{OOD}
\def\obsspace{\mathcal{O}}
\def\actspace{\mathcal{A}}
\def\detspace{\mathcal{B}}
\def\locspace{\mathcal{M}}
\def\detelem{b}
\def\locelem{M}
\def\navpolicy{\pi_{\text{IL-OOD}}}
\def\recpolicy{\pi_{\text{Rec}}}
\def\regpolicy{\pi_{\text{IL}}}
\def\dataset{\mathcal{D}}
\def\calibset{\mathcal{C}}
\def\predband{C_{\alpha}}

\def\endrecthresh{t_{\text{max\_tries}}}

\def\normflow{NF}
\def\rnd{RND}
\def\autoenc{AE}
\def\vaerecon{VAE-R}
\def\vaekl{VAE-KL}

\def\weblink{\urllink[pre = \bgroup\bf, post = \egroup]}

\makeatletter
\let\NAT@parse\undefined
\makeatother
\usepackage{hyperref}

\title{\textbf{\fr{}}: \underline{In}formed \underline{F}ailur\underline{e} \underline{R}esilience in Learned Visual Navigation Control}

\author{Zishuo Wang$^{1}$, Joel Loo$^{1}$, and David Hsu$^{1}$
\thanks{$^{1}$School of Computing \& Smart Systems Institute, National University of
Singapore. {\tt\small \{zishuo.wang, joell, dyhsu\}@comp.nus.edu.sg}}
}%


\begin{document}

\maketitle
\thispagestyle{empty}
\pagestyle{empty}

\begin{abstract}


While imitation learning (IL) has enabled successful visual navigation in many common environments, IL policies are prone to unpredictable failures under out-of-distribution (\ood{}) scenarios. This necessitates \textit{failure-resilient} policies, which not only detect failures, but also recognise their sources and  recover from them autonomously. We propose \textit{\fr{}}, a general framework for building IL policies with \textit{informed failure resilience} without failure or recovery demonstrations.  
Given an IL policy's architecture and training dataset, \fr{} retrains a new \ood{}-aware policy with a latent space structured for \ood{} failure detection, with a modified objective based on the Variational Information Bottleneck. 
It applies a visual explainability technique, Grad-CAM, to attribute failures to specific image regions and inform a heuristic policy for recovery.  
All these are achieved without requiring additional training data. 
Real-world experiments show that \fr{} enables informed failure recovery across two different policy architectures, supporting long-range navigation in complex environments. 

\end{abstract}

\section{Introduction}

Visual navigation is an attractive approach to robot navigation, leveraging rich visual information from low-cost sensors~\cite{boninfont2008survey}. Imitation learning (IL) has emerged as a key method to learn visual navigation policies~\cite{ai2022deep, shah2023gnm, sridhar2024nomad}, but is inherently limited by training data. IL policies may fail unpredictably on inputs outside the training distribution, often without clear explanation~\cite{chalapathy2019deep, richter2017safe, wong2022momart}. We consider the problem of building IL policies with \textbf{informed failure resilience}, which we argue extends beyond detecting failures to also encompass autonomous, informed recovery from failures.

Current works addressing failure have two key limitations. First, they focus on detecting failures, with handling often limited to triggering operator interventions~\cite{gokmen2023help, xu2025can}. Second, they often rely on external failure detectors, which lack access to policies' internal representations and distributions and may fail to capture their true failure modes~\cite{mcallister2019robustness}. These limitations motivate three desiderata for failure-resilient policies: \textbf{(i)} the ability to autonomously \textit{recover} from failures; \textbf{(ii)} the ability to \textit{recognise} failure sources to inform recovery; and \textbf{(iii)} detection and recognition grounded in the policy's internal representations to capture its true failure modes.

We propose \textbf{\fr{}}, a general framework for building informed, failure-resilient visual navigation policies \textit{without any failure or recovery demonstrations}. \fr{} operates within the standard offline IL setting, learning policies from a single, static dataset of successful demonstrations, where failures are naturally approximated as scenarios outside the dataset's distribution~\cite{agia2024sentinel, richter2017safe, ccatal2020anomaly, wong2022momart}. Given this dataset and a specific policy architecture, \fr{} retrains from scratch an \ood{}-aware policy capable of detecting and recognising out-of-distribution (\ood{}) elements in its inputs, without any additional data. \fr{} pairs it with a recovery policy guided by \ood{} recognition to recover from \ood{} scenarios. Learning a recovery policy is ill-posed since failure and recovery examples are absent from data. Instead, \fr{} designs a heuristic to greedily distance the robot from the identified failure, then perturb it back to a nominal, in-distribution state.

\begin{figure}[!t]
  \includegraphics[width=\linewidth]{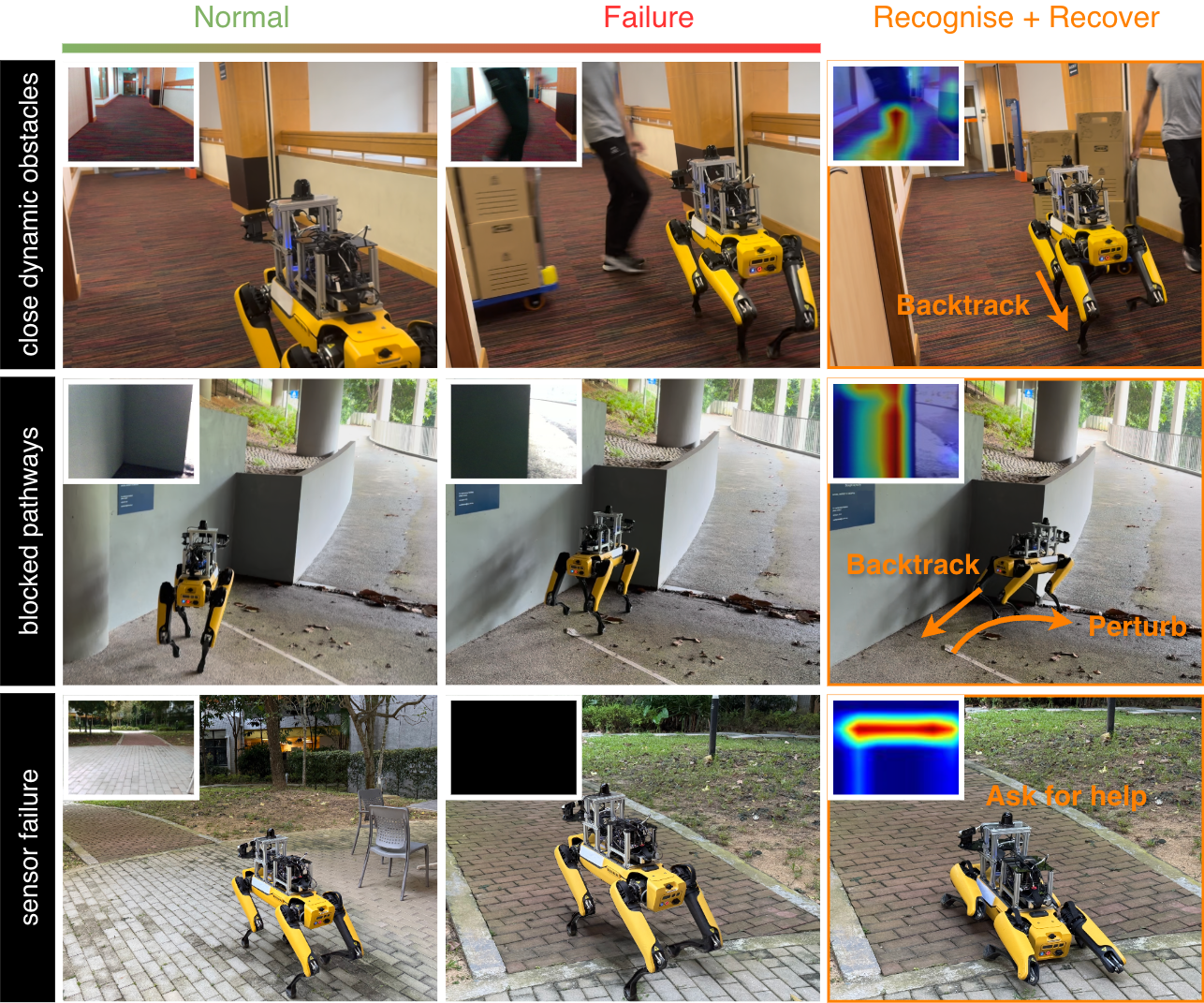}
  \caption{\textbf{\fr{} enables autonomous recovery from policy failures.} Limited training data causes failures in \ood{} scenarios (\eg{} sensor failures, close dynamic obstacles \etc{}). \fr{} \textit{detects} these, and further \textit{recognises} the associated \ood{} image regions, enabling \textit{informed failure recovery}.}
  \label{fig:splash}
  \vspace{-2.0em}
\end{figure}

The \ood{}-aware policy is retrained using a Variational Information Bottleneck (VIB) objective, which structures the latent space between the vision encoder and action decoder to distinguish in- and out-of-distribution inputs. The resulting representation enables us to compute an \ood{} detection score and apply Grad-CAM explainability methods to identify image regions that contribute to the detection~\cite{selvaraju2019gradcam, leem2024attention}. Importantly, \fr{} does not change the policy's original encoder or decoder, adding only two lightweight linear layers to parameterise the stochastic latent representation. As shown in \autoref{fig:splash}, \fr{} handles diverse failure modes by detecting and recognising \ood{} elements (\eg{} dynamic obstacles, close-up obstructions, and sensor failures), then performs recovery informed in real time by the recognition feedback. We show that \fr{} enables two policies---DECISION~\cite{ai2022deep} and GNM~\cite{shah2023gnm}---to detect and recover from critical failures in the real world, allowing a robot to complete a challenging 300m indoor--outdoor route without intervention.

\section{Related Work}
\label{sec:related_work}

\subsection{Learned visual navigation} 

Navigating purely with high-dimensional image inputs is challenging for classical robotic systems~\cite{boninfont2008survey}. Recent progress in visual navigation is enabled by learning perceptual modules~\cite{frey2023wvn, meng2023terrainnet} or even entire policies end-to-end~\cite{ai2022deep, shah2023gnm, sridhar2024nomad}. Among these, end-to-end imitation learning (IL) has proven especially effective~\cite{ai2022deep, shah2023gnm, sridhar2024nomad}, enabling robust navigation in the wild~\cite{gao2024intentionnet, shah2022viking}. However, such policies remain opaque and offer no indication of impending failure~\cite{mcallister2019robustness}. We address this by defining \textit{informed failure resilience} for end-to-end IL policies and proposing \fr{}, a framework to augment IL policies into failure-resilient policies.

\subsection{Failure detection} 


Robust visual navigation in open-world settings hinges on the ability to detect failures~\cite{rahman2021monitoringsurvey}. Some recent approaches learn from explicit failure data: BADGR~\cite{kahn2021badgr} heuristically labels failures in collected training data post hoc, while others learn from human interventions that precede failures~\cite{spencer2020interventions, liu2023modelbased, liu2024siriusfleet}. Notably, \cite{liu2024siriusfleet} combines this with novelty detection to capture both user-defined failures and \ood{} events.

Since imitation learning (IL) datasets typically include only successful trajectories, unsupervised failure detection without explicit failure data becomes crucial when learning policies with IL. Failure detection is often framed as identifying \ood{} inputs that lie outside the IL dataset's distribution~\cite{richter2017safe}. This can be approached via epistemic uncertainty estimation~\cite{itkina2022interpretable} using ensembles or Bayesian neural networks~\cite{mcallister2019robustness}, but such methods are often too computationally intensive for real-time navigation with limited compute~\cite{xu2025can}.

Alternative methods seek greater efficiency, often at the cost of using detectors external to the policy which may not fully capture its failure modes and tend toward overly conservative detections~\cite{mcallister2019robustness}. Autoencoders~\cite{richter2017safe, wong2022momart} and density estimators~\cite{wellhausen2020safe, cai2024evora} are common approaches, but may not accurately capture the data distribution as they are trained on observations and exclude action data, often producing overly conservative detections~\cite{mcallister2019robustness}. \cite{he2024rediffuser} apply Random Network Distillation to learn an external detector for the joint observation-action distribution. FAIL-Detect~\cite{xu2025can} compares a range of \ood{} scoring approaches, and combines them with conformal prediction for \ood{} detection with statistical guarantees. Finally, some methods leverage LLMs to identify and explain failures~\cite{sinha2024rtanomaly, jin2025logicad}, though these lack efficiency and insight into the policy's specific failure modes.

We emphasise that efficiency and capturing a policy's true failure modes are key needs. \fr{} attempts to address these by embedding failure detection within the IL policy.

\subsection{Failure recovery}

Robust navigation requires not only the ability to detect failures, but also to independently recover from them. Existing approaches largely terminate the episode on detection, and seek human intervention~\cite{hecker2018failure, mcallister2019robustness, gokmen2023help}. A small subset of works attempts to autonomously recover from failure by executing predefined actions open-loop---\eg{} returning to a default state~\cite{wong2022momart} or executing a default behaviour~\cite{richter2017safe}. Some systems layer increasingly aggressive preset actions~\cite{macenski2020marathon2} to improve recovery chances. \cite{del2018not} further requests interventions and learns from subsequent recovery demonstrations.

\fr{} emphasises robustness through autonomous recovery and highlights the importance of \ood{} recognition in providing cues for informed, efficient recovery. To build failure-resilient policies, \fr{} combines \ood-aware policies with unified \ood{} detection and recognition embedded inside, with a recovery policy guided by recognition results.

\section{Problem formulation}
\label{sec:formulation}

IL for visual navigation learns a policy $\regpolicy{}: \obsspace{}\rightarrow\actspace{}$ from a dataset $\dataset{} = \{(o_1, a_1), \dots, (o_N, a_N)\}$, where $o_i\in\obsspace{}$ are RGB or depth images, and $a_i\in\actspace{}$ are actions typically given as target velocities. An \ood-aware policy is $\navpolicy: \obsspace\rightarrow\actspace\times\detspace\times\locspace$, mapping observation $o_t$ to action $a_t$, binary \ood{} detection $\detelem_t\in\detspace$ and a representation of the location of \ood{} regions influencing the detection, $\locelem_t\in\locspace$. We seek to obtain $\navpolicy{}$, given a specific policy network architecture and $\dataset{}$. Since many IL datasets only contain successful demonstrations, $\navpolicy{}$ must be obtained without requiring additional failure data. We also seek to obtain $\recpolicy{}$, a recovery policy which modulates or overrides $\navpolicy{}$ to take corrective actions in \ood{} scenarios, selecting appropriate actions on the basis of spatial feedback from \ood{} recognition $\locelem_t$. $\recpolicy{}$'s actions either recover the robot to an in-distribution state, or halt it and call for external assistance if recovery is not possible. $\navpolicy{}$ and $\recpolicy{}$ together constitute a failure-resilient policy.

\section{Approach}

\begin{figure*}[!t]
\centering
  \includegraphics[width=\linewidth]
  {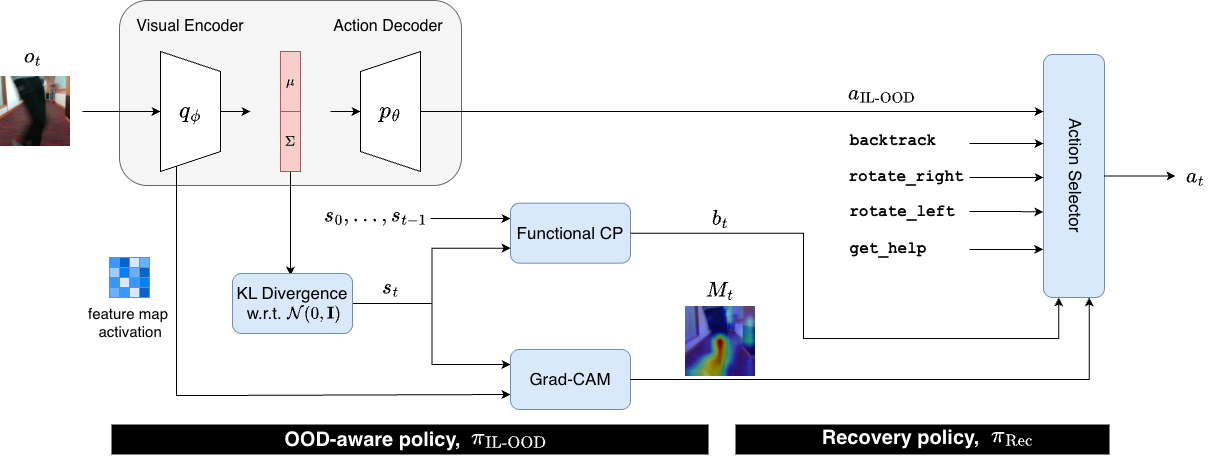}
  \caption{\textbf{\fr{} framework for informed failure resilience.} \fr{} achieves informed failure resilience through augmenting learned policies with \ood{}-awareness. Apart from generating actions, \ood{}-aware policies also detect when \ood{} inputs are encountered ($\detelem{}_t$) and localise the \ood{} features in the observations ($\locelem{}_t$). Detection and recognition enable informed recovery from \ood{} or failure scenarios. \fr{} employs a heuristic recovery policy taking $\detelem{}_t, \locelem{}_t$ as inputs.}
  \label{fig:system}
  \vspace{-1.0em}
\end{figure*}

\textbf{\fr{}} (\autoref{fig:system}) provides a framework to modify visual navigation policies trained with IL, $\regpolicy{}$, into \ood{}-aware policies, $\navpolicy$. It also provides a heuristic recovery policy, $\recpolicy{}$, that makes corrective actions in \ood{} scenarios based on \ood{} recognition information from $\navpolicy{}$.

\subsection{\ood{}-aware learned navigation policy, \texorpdfstring{$\navpolicy{}$}{}}



\ood{}-awareness involves detecting \ood{} scenarios and pinpointing the corresponding \ood{} regions in the visual input. We emphasise that \ood{}-awareness should be task-relevant, identifying \ood{} scenarios and pinpointing \ood{} regions related to failures on the navigation task the policy is trained for. Our policy, $\navpolicy{}$, takes an image input $I_t \in \mathbb{R}^{3 \times H \times W}$ and outputs action $a_t$, a binary \ood{} flag $\detelem_t$, and a heatmap $\locelem_t \in \mathbb{R}^{H \times W}$ that identifies the \ood{} features in $I_t$. \fr{} augments a policy $\regpolicy{}$ into an \ood{}-aware one in two phases. First, it reinitialises the policy with minor modifications for its latent space to represent a diagonal Gaussian, then retrains it with a regularised behaviour cloning objective to yield a structured latent space capturing task-relevant factors from $I_t$. Second, during inference, conformal prediction and Grad-CAM are applied to features from the policy's latent space for \ood{} detection and recognition respectively.

\textbf{Learning task-relevant \ood{}-awareness with VIB.} We learn task-relevant latent representations $Z$, with the Variational Information Bottleneck~\cite{alemi2017vib}. This compresses input observations $O$ into $Z$, while maximally retaining information crucial to predicting output actions $A$. Given $I(\cdot ;\cdot)$ as mutual information, we maximise:

\begin{equation}
    \mathcal{L}_{\text{VIB}} = I(A; Z) - \beta I(Z; O)
\end{equation}

Following \cite{alemi2017vib}, we apply variational approximations to optimise this intractable objective, yielding the objective:

\begin{equation}
    \max_{\phi, \theta} \frac{1}{|\mathcal{D}|} \sum_{(o, a)\in\mathcal{D}} \mathbb{E}_{q_\phi(z|o)}[\log p_\theta(a|z)] - \beta\text{KL}[q_\phi(z|o) \lVert p(z)]
\end{equation}
where $q_\phi, p_\theta$ are the visual encoder and action decoder, respectively. The first term simplifies to a behaviour cloning objective, while the second term regularises the latent representation $Z$ to stay close to prior $p(z)$, which we define as a unit Gaussian $\mathcal{N}(0, \textbf{I})$. This regularisation means that at inference, an \ood{} input $\hat{o}$ is likely to produce latent feature $\hat{z}$ that diverges significantly from the prior. Thus the KL divergence $\text{KL}[q_\phi(z|o) \lVert \mathcal{N}(0, \textbf{I})]$ can serve as an effective scalar score indicative of \ood{} inputs during inference.

Learning with VIB can be viewed as augmenting a standard IL setup with additional regularisation of the latent features between a policy's visual encoder and action decoder. Architecturally, this requires only the addition of two lightweight linear layers to represent mean and variance, with the original encoder and decoder structure unchanged. This is a lightweight approach to instil task-relevant \ood{}-awareness across a range of IL policies, without supervision using explicit failure data.

\textbf{\ood{} detection with conformal prediction.} 
Binary detection $\detelem_t$ involves thresholding scalar \ood{} scores, but setting appropriate thresholds is challenging, especially since the threshold may be time-varying due to changing dynamics and scene appearance~\cite{xu2025can}. Conformal prediction (CP), particularly functional CP, offers a principled, distribution-free way to compute thresholds by generating a prediction band $\predband{}$ at a user-defined significance level $\alpha \in [0, 1]$. Under mild assumptions, $\predband{}$ contains the true score with probability $1 - \alpha$, enabling confident rejection of scores outside the band~\cite{diquigiovanni2025functionalcp, xu2025can}. Functional CP is an efficient, training-free method that wraps around a sequence of scores from any model.

We apply functional CP to a calibration set of trajectories to obtain time-varying thresholds on the KL scores. Similar to \cite{xu2025can}, we predict a one-sided $\predband{}$ that provides an upper bound on whether a score is in-distribution with probability $1-\alpha$. We calibrate on a held-out split from the IL dataset, $\calibset{}$. We chunk trajectories into non-overlapping segments of $T$ steps to standardise length, obtaining $\calibset{}_{\text{scores}} = \{(s_{i, 1},\dots,s_{i,T}): i=1,\dots,\lvert\calibset{}\rvert\}$. Following \cite{diquigiovanni2025functionalcp}, we split $\calibset{}$ into $\calibset{}_{\mu}$ to compute the mean score across trajectories for each timestep, and $\calibset_{w}$ to compute the width of the band $\predband{}$. Given the mean scores $\mu = (s_{1},\dots,s_{T})$, $w$ is computed based on the maximum deviations from $\mu$ per trajectory. Specifically, given
\begin{equation}
S = \{d_j: j=1,\dots,\lvert\calibset{}_{w}\rvert\} \text{ where } d_j = \max\limits_{t\in[1, T]} \{s_{j, t} - \mu_t\},
\end{equation}
$w$ is the $(1-\alpha)$-quantile of $S$. The prediction band is the set of intervals $\{[-\infty, \mu_t + w]: t=1,\dots,T\}$, where scores above the upper bound are rejected as \ood{}. \hyperref[app:coverage]{Appendix~\ref*{app:coverage}} provides a brief discussion of how our procedure ensures coverage.

\textbf{\ood{} recognition with Grad-CAM.} Recognition aims to identify regions in observations $o\in\obsspace{}$ which contribute to the \ood{} detection. To do so, we apply Grad-CAM~\cite{selvaraju2019gradcam}, a visual explanation method for interpreting CNN outputs. Grad-CAM is a training-free module integrated into $\navpolicy{}$ at test time, offering efficiency as it requires only a single backward pass through the visual encoder.

Grad-CAM produces recognition heatmaps that highlight spatial regions influencing the model's predictions. It computes gradients of the prediction with respect to feature map activations in preceding convolutional layers, obtaining weights capturing the contribution of each feature map to the prediction. To localise \ood{}-related visual features, we apply Grad-CAM on the KL-based \ood{} scores predicted by the model. As deeper layers tend to capture higher-level semantic concepts (\eg{} object parts)~\cite{mahendran2016visualize}, we use Grad-CAM on the last layer of the visual encoder $q_\phi$ to encourage meaningful recognition results. Specifically, we compute the weight for the $k$-th feature map of the layer as:
\begin{equation}
    \alpha_k = \frac{1}{N}\sum\limits_i\sum\limits_j \frac{\partial \text{KL}[q_\phi(z_t|o_t) \lVert \mathcal{N}(0, \textbf{I})]}{\partial A^{k}_{ij}}
\end{equation}
where $A^{k}$ is the $k$-th feature map and $N$ normalises across all $A^{k}$. The recognition result is:
\begin{equation}
    M_{\text{Grad-CAM}} = \text{ReLU}\left( \sum_{k} \alpha_k A^{k} \right)
\end{equation}
The resulting $M_{\text{Grad-CAM}}$ has the same dimensions as $A^{k}$, which we upsample with bilinear interpolation to obtain $M$.

While we formulate the \fr{} system using this vanilla Grad-CAM version, many improved variants of Grad-CAM~\cite{chattopadhyay2018gradcampp, leem2024attention} exist, which can directly replace Grad-CAM in \fr{}. We demonstrate this in \hyperref[app:gradcampp]{Appendix~\ref*{app:gradcampp}}, by integrating an extension of Grad-CAM which enables improved recognition performance.

\subsection{Heuristic recovery policy, \texorpdfstring{$\recpolicy{}$}{}}
\label{sec:approach_recovery}

The recovery policy, $\recpolicy{}$, aims to take informed corrective actions when \ood{} scenarios are detected, and otherwise pass through $\navpolicy{}$'s actions during normal operation. 
The policy uses \ood{} recognition to select an appropriate recovery action.
We design a set of recovery action primitives aimed at enabling the robot to distance and orient itself away from identified \ood{} regions in $o_t$.
Specifically, these actions \textbf{(i)} \emph{backtrack} the robot through known safe states, \textbf{(ii)} \emph{perturb} it out of the \ood{} scenario, \textbf{(iii)} halt the robot and \emph{seek human assistance}, if the robot is unable to recover before some termination condition is reached.

Concretely, $\recpolicy{}$ has action set \{\texttt{backtrack}, \texttt{rotate\_left}, \texttt{rotate\_right}, \texttt{get\_help}\}. During recovery, it selects among the first three actions; \texttt{get\_help} is invoked only after $\endrecthresh{}$ unsuccessful recovery attempts, when $\detelem{}_t$ has not returned to a non-\ood{} state. 
To select an action, we divide the heatmap $\locelem{}_t$ into three vertical bins---left, centre, and right---which aligns with the discretisation of our recovery actions. A bin is marked as \ood{} if its summed pixel values exceed a threshold. $\recpolicy{}$ then applies simple heuristics to map the bins to \ood{} actions: if the right (left) bin is \ood{}, it selects \texttt{rotate\_left} (\texttt{rotate\_right}) to steer away from the detected region; otherwise, it selects \texttt{backtrack}, which replays a cached sequence of past actions in reverse.



\section{Experiments}
\label{sec:result}

We evaluate three hypotheses: \textbf{(H1)} \fr{} produces \ood{}-aware policies that are effective at task-relevant \ood{} detection and recognition. \textbf{(H2)} Our \fr{} policy is effective at recovering from failures of the original visual navigation policy. \textbf{(H3)} \fr{} is applicable across different visual navigation policies. 

\autoref{sec:exp_det_rec} validates \textbf{H1} by comparing with selected baselines on a test set comprising RGB data from 3 Realsense D435i cameras, collectively covering a 140$^\circ$ horizontal FoV. \autoref{sec:exp_fr_nav} validates \textbf{H2} and \textbf{H3} through real-world experiments on a Boston Dynamics Spot robot with the above Realsense setup, with policies running onboard an Nvidia Orin AGX. 
Finally, we show that \fr{} can improve navigation with learned controllers, enabling effective failure recovery and thereby successful completion of a long-range route~(\autoref{fig:long_range_results}).

We train our policies and baselines on the training dataset from \cite{ai2022deep}, comprising 30 hours of teleoperated trajectories demonstrating the task of path-following navigation in indoor and outdoor scenes. We identify four types of Navigation Failures (NFs) that are \ood{} for this dataset, spanning recoverable failures and irrecoverable ones for which the correct response is to request human assistance.

\begin{enumerate}[label=\textbf{NF\arabic*}]
    \item \label{failure:sensor} \textbf{Sensor failure.} The robot receives blacked out images. The expected outcome is to seek human assistance, as the robot cannot recover on its own.
    \item \label{failure:blocked} \textbf{Blocked pathways.} The robot's path is blocked by obstacles or untraversable regions. Since the dataset's trajectories focus on following clear paths, this is \ood{}. We consider \textbf{(i)} \textit{local minima}, from which recovery is possible through a short sequence of local, reactive actions, and also \textbf{(ii)} \textit{dead-ends}, which require substantially longer interventions or replanning to escape, and are irrecoverable for a local navigation policy.
    \item \label{failure:dyn} \textbf{Close-range dynamic obstacles.} Sudden dynamic obstacles appearing close to the robot. As successful demonstrations avoid dynamic obstacles at a distance, such situations are \ood{}. The expected outcome is to halt and backtrack/perturb away from the sudden obstacle, then resume navigation along the path.
    \item \label{failure:goal} \textbf{Infeasible image goals.} Specific to image goal-conditioned policies where the goal is dissimilar to current observations. We focus on the irrecoverable case where the goal is not locally reachable, and the expected outcome is for the robot to seek human assistance.

\end{enumerate}



\subsection{\ood{} detection and recognition}\label{sec:exp_det_rec}
\begin{table}[!t]
\caption{\textbf{Selected baselines for \ood{} detection/recognition.}}
\vspace{-2em}
\scriptsize
    \begin{center}
        \begin{tabular}{@{\extracolsep{4pt}}ccccc@{}}
            \toprule
            \textbf{Type} & \textbf{Method} & \textbf{Det.} & \textbf{Loc.} & \textbf{Inputs} \\
            \midrule
            \textit{Density} & \textit{\normflow{}}~\cite{wellhausen2020safe} & Density & - & $O_t$ \\
            \cmidrule{1-5}
            \textit{One-class} & \textit{\rnd{}}~\cite{he2024rediffuser} & Novelty & - & $O_t, A_t$ \\
            \cmidrule{1-5}
            \multirow{3}{*}{\textit{Recon}} & \textit{\autoenc{}}~\cite{richter2017safe} & Recon & Recon & $O_t$ \\
            & \textit{\vaerecon{}} & Recon & Recon & $O_t$ \\
            & \textit{\vaekl{}} & KL & Grad-CAM & $O_t$ \\
            \cmidrule{1-5}
            \multirow{2}{*}{\textit{\fr{} (BC)}} & \textit{\frinet{}} & KL & Grad-CAM & $O_t, A_t$ \\
            & \textit{\frdec{}} & KL & Grad-CAM & $O_{t-H},\dots,O_t, A_t$ \\
            \bottomrule
        \end{tabular}
    \label{tab:baselines}
    \end{center}
    \vspace{-2.5em}
\end{table}

\begin{figure*}[!t]
  \includegraphics[width=\linewidth]{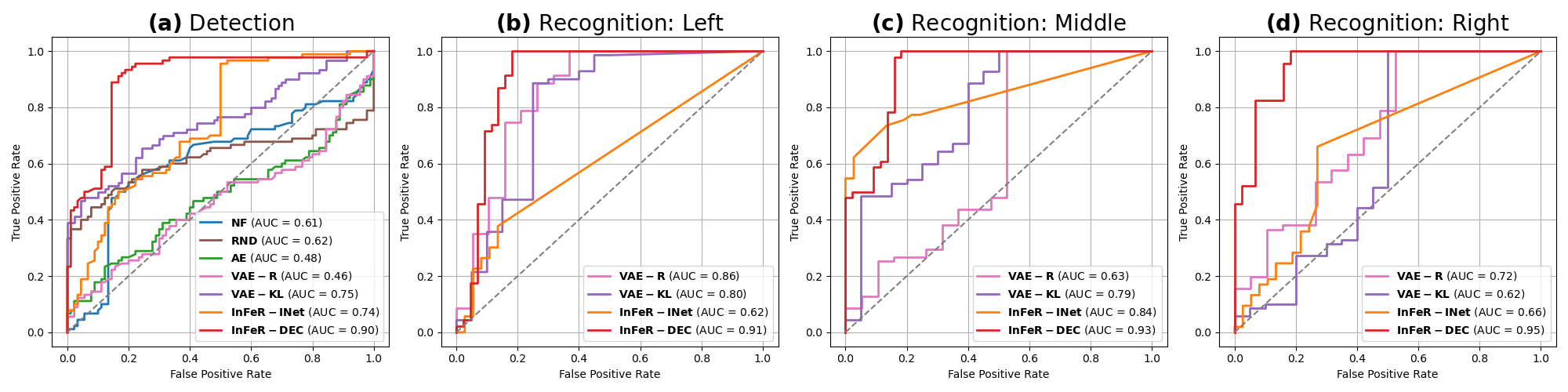}
  \centering
  \caption{\textbf{Comparison to baselines for \ood{} detection and recognition.} (\textit{a}) ROC curves for \ood{} detection across CP significance levels $\alpha$. (\textit{b-d}): ROC curves for \ood{} recognition, based on scores from summing pixels across the heatmap. We compute scores per heatmap bin: \textit{Left}, \textit{Middle}, \textit{Right}.}
  \label{fig:baselines}
\end{figure*}

\begin{figure*}[!t]
  \includegraphics[width=\linewidth]{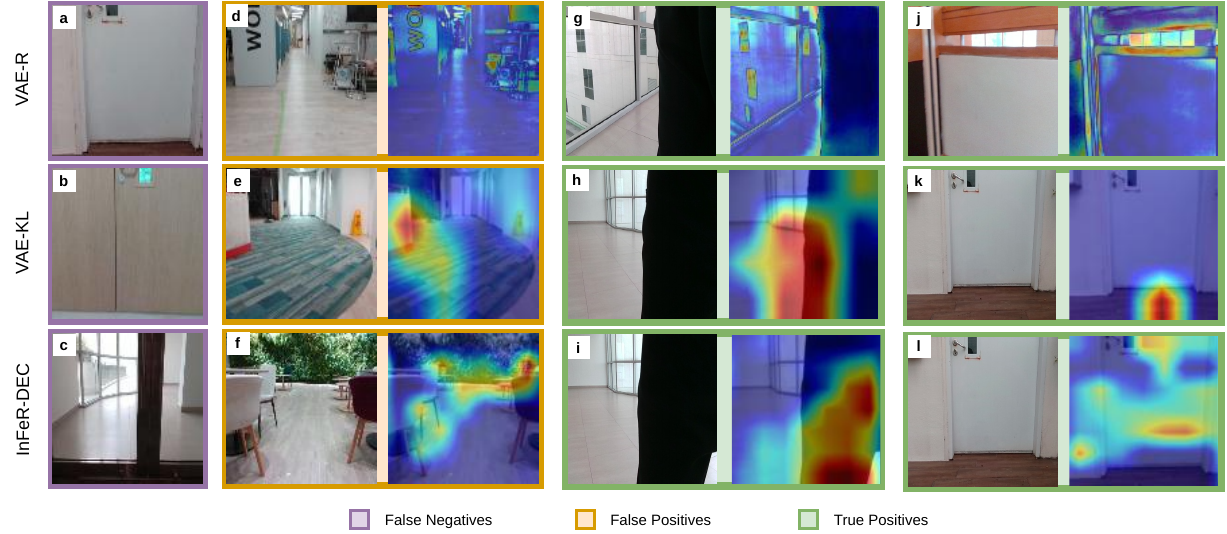}
  \centering
  \caption{\textbf{\ood{} detection and recognition examples.} We visualise selected reconstruction and \fr{} methods. (\textit{a-c}): Reconstruction-based methods struggle to detect failures like blocked pathways when there are few visual features. (\textit{d-f}): \vaerecon{} indiscriminately highlights high-frequency features, falsely detecting failures. \frdec{} is sensitive to obstacles (\eg{} chairs, tables), prematurely triggering a detection. (\textit{g-l}): \vaekl{} offers smoother detections than \vaerecon{}, but may not identify areas directly impacting navigation. In contrast, \frdec{} correctly highlights the agent's leg (\textit{i}) and the door (\textit{l}) as failure causes.} 
  \label{fig:det_loc_qualitative} 
  \vspace{-1.0em}
  
\end{figure*}


We evaluate performance of \ood{}-aware policies on \ood{} detection and recognition, comparing them against selected baselines (\autoref{tab:baselines}). The \ood{}-aware policies are based on INet~\cite{gao2024intentionnet} and DECISION~\cite{ai2022deep}. The baselines belong to three common classes: \textit{density-based} approaches learn a density estimator for the training distribution; \textit{one-class discriminators} learn to output a continuous metric without directly modelling the training distribution; \textit{reconstruction-based} approaches indirectly model the training distribution by learning to reproduce observations, with post-hoc metrics applied to detect and recognise \ood{} inputs.


For density-based methods, we fit a normalising flow (\normflow{}) to observations with a Gaussian prior, using estimated likelihoods as the \ood{} score. For one-class discriminators, we use Random Network Distillation (\rnd{})~\cite{he2024rediffuser}, which trains a predictor to map training data to output of a fixed random function, using prediction error as a measure of sample novelty and the \ood{} score. In prior art, these two approaches detect \ood{} inputs but do not support recognition. Reconstruction-based methods include autoencoders (\autoenc{})~\cite{richter2017safe} and variational autoencoders (\vaerecon{})~\cite{wong2022momart}. We apply post-hoc metrics based on reconstruction error: mean-squared error for \ood{} detection, and pixel-wise errors for recognition. Since \fr{}'s detection/recognition metrics apply to any VIB-based model, we test them with \vaekl{}, which shares weights with \vaerecon{}, but uses KL scores and Grad-CAM for \ood{} detection and recognition instead.





We collect a test set of 180 trajectories: 90 with a single failure near the end, and 90 normal (\autoref{fig:splash}). Failures are evenly distributed across three types---sensor failure, blocked paths, and close-range dynamic obstacles (\ref{failure:sensor}-\ref{failure:dyn}). We exclude infeasible image goals, which is a failure specific to image goal-conditioned architectures. Half the trajectories are in novel office and garden environments; the rest are in novel areas that resemble training data. Each trajectory is 10s long, with failures occurring at the 8s mark. A failure is detected if any frame from 8s onward is classified as \ood{}. To annotate ground-truth recognition, we divide each frame vertically into three bins and mark the bins containing features related to navigation failure (\eg{} dynamic objects, blockages), similar to the approach in \autoref{sec:approach_recovery}.


We illustrate OOD detection performance via ROC curves across CP significance levels $\alpha$. To evaluate recognition independently of detection, we restrict analysis to failure trajectories and, for each method, to the detections obtained at a true positive rate of 0.9, normalising detection sensitivity across methods. On this subset, we compute per-bin ROC curves for the recognition heatmap, with scores given by $\mathbb{I}[b_t = 1] \sum_{m \in M_{t,k}} m$, where $b_t = 1$ is an \ood{} detection, and $M_{t,k}$ is the $k$-th bin of the heatmap $M_t$. This yields the per-bin recognition score only when the input is flagged as OOD. The results in \autoref{fig:baselines} support \textbf{H1}: \fr{} produces \ood{}-aware policies effective at task-relevant \ood{} detection/recognition. We draw three conclusions:



\textbf{\ood{}-aware policies outperform existing methods.}
\fr{}'s integration of \ood{} detectors into IL policies outperforms state-of-the-art detection and recognition approaches, supporting \textbf{H1}. Both \fr{} methods achieve high AUC scores across \textit{both} detection and recognition compared with density-based approaches (\normflow{}), one-class discriminators (\rnd{}), and reconstruction-based approaches (\autoenc{}, \vaerecon{}). \frdec{} achieves the best performance by a clear margin, with high TPR and relatively few false positives. It outperforms \frinet{}, likely because its temporally accumulated features offer richer information for distinguishing \ood{} scenarios. Unlike other methods that detect and recognise independently of the policy (\eg{} \vaerecon{}, \vaekl{}), \fr{} embeds both in the policy, reducing computation.

\textbf{KL scores, Grad-CAM are most effective metrics.} 
Our results suggest that using KL scores for detection and Grad-CAM for recognition offers better performance than the mean-squared and pixel-wise errors typical in \textit{Recon} methods. Specifically, \vaekl{} achieves stronger performance on both tasks despite having the same model weights as \vaerecon{}, differing only in the \ood{} measures used.
We hypothesise that this is due to KL scores and Grad-CAM operating on \textit{latent vectors} instead of high-dimensional reconstructed images. As latent vectors are compressed representations that learn to focus on task-relevant factors, they are likely less noisy information sources than reconstructed images. Qualitative analysis (\autoref{fig:det_loc_qualitative}) bears this out: recognition results from \vaerecon{} are significantly noisier than \vaekl{} or \fr{} methods, and overfit to high-frequency artefacts in the inputs. While recent works shift the focus to methods like density-based~\cite{wellhausen2020safe} or second-order~\cite{ancha2024evidential, xu2025can} approaches, we highlight that selecting the right \ood{} measures enables reconstruction methods to outperform these on detection. This also validates \fr{}'s choice of scores for \ood{} detection and recognition.

\textbf{Action supervision improves \ood{} recognition.} We hypothesised that learning from both observations and actions encourages more task-relevant representations. Observation-only methods (\eg{} \textit{Recon}) perform comparably on detection. However, adding actions enables better, task-relevant \ood{} recognition: in \autoref{fig:det_loc_qualitative} (\textit{k, l}), \frdec{} highlights features relevant to the path-following task (a closed door), whereas \vaekl{} highlights irrelevant textures on the floor.









\subsection{Failure resilient navigation in the real world}\label{sec:exp_fr_nav}

We evaluate \fr{}’s effectiveness in enabling failure-resilient navigation in the real world, across different policy architectures. Specifically, we first compare the performance of the base policy with that of the \fr{} variant in terms of the original navigation capability captured by the training set. Then we test whether \fr{} policies can detect and recover from the key failure types identified (\ref{failure:sensor}–\ref{failure:goal}). To illustrate the importance of informed failure resilience, we show that \fr{} allows the DECISION policy to autonomously complete a 300m route with varied terrain (cluttered indoor plaza, outdoor park with forested paths) despite critical failures.

We test two policies with differing architectures: \frgnm{}, based on the General Navigation Model~\cite{shah2023gnm}, an image goal-conditioned policy using CNN encoders; \frdec{}, based on DECISION~\cite{ai2022deep}, a recurrent neural network with a more complex structure to explicitly handle multimodality in navigation tasks. 
We also validate that application of the InFeR framework to a policy does not adversely affect the policy's base performance on in-distribution inputs. We follow the metrics and tasks used in \cite{ai2022deep}. \autoref{tab:VIBcompare} shows only a minimal reduction in performance for \frdec{} compared to DECISION.

\begin{table}[!t]
\caption{\textbf{Policy performance with and without VIB.} }
\centering
\fontsize{9}{12}\selectfont
\begin{tabular*}{\linewidth}{@{\extracolsep{\fill}}ccccc@{}}
\hline
                & \multicolumn{2}{c}{Task i} 
                & \multicolumn{2}{c}{Task ii} \\
\cline{2-5}
\textbf{Method} & \textbf{SR}    & \textbf{CR}    & \textbf{SR}  & \textbf{CR}   \\
\hline
DECISION        & 0.9   & 0.988 & 0.9 & 0.98 \\
\frdec{}       & 0.9   & 0.98  & 0.8 & 0.96 \\
\hline
\end{tabular*}
\label{tab:VIBcompare}
\vspace{-1.0em}
\end{table}


To test failure detection and handling performance, \frgnm{} is evaluated on all failure types except blocked pathways~(\ref{failure:blocked}), since dead-ends/local minima occlude the goal, making this failure type similar to infeasible goals~(\ref{failure:goal}). Conversely, since \frdec{} learns path-following skills and has no goal conditioning, it is tested on all failure types except infeasible goals. For each failure type, we run 20 trials, triggering a failure after $\sim$8s of normal operation. We report detection success rate (Det. SR), combined detection and recovery success rate (Han. SR), and time taken from first detection to recovery. We draw three conclusions:

\begin{table*}[!t]
\caption{\textbf{Evaluation of \fr{} policies.} 
\fr{} is applied to GNM~\cite{shah2023gnm} and DECISION~\cite{ai2022deep}. 
A ``blind'' variant that does not use recognition results, and randomly selects 
corrective actions on \ood{} detections is also evaluated.}
\centering

\resizebox{\textwidth}{!}{%
\begin{tabular}{@{}ccccccccccc@{}}
    \toprule
    \textbf{} & \textbf{} 
    & \multicolumn{3}{c}{\frgnm{}}
    & \multicolumn{3}{c}{\frdec{}}
    & \multicolumn{3}{c}{\undec{}} \\
    
    \cmidrule(lr){3-5}
    \cmidrule(lr){6-8}
    \cmidrule(lr){9-11}
    
    \textbf{Failure type} 
    & \textbf{Recoverable} 
    & \textbf{Det. SR} 
    & \textbf{Han. SR} 
    & \textbf{Time (s)} 
    & \textbf{Det. SR} 
    & \textbf{Han. SR} 
    & \textbf{Time (s)} 
    & \textbf{Det. SR} 
    & \textbf{Han. SR} 
    & \textbf{Time (s)} \\
    
    \midrule
    
    Sensor failure 
    & \xmark 
    & 1 & 1 & 10.00$\pm$0.00 
    & 1 & 1 & 5.00$\pm$0.00 
    & 1 & 1 & 5.00$\pm$0.00 \\
    
    Blocked (Dead-end) 
    & \xmark 
    & - & - & - 
    & 1 & 0.95 & 11.33$\pm$6.90 
    & 1 & 0.75 & 19.83$\pm$6.04 \\
    
    Blocked (Local minima) 
    & \cmark 
    & - & - & - 
    & 1 & 0.75 & 7.32$\pm$2.42 
    & 1 & 0.55 & 12.11$\pm$3.46 \\
    
    Close-range dynamic obs. 
    & \cmark 
    & 0.9 & 0.7 & 8.64$\pm$2.62 
    & 0.9 & 0.85 & 2.19$\pm$1.56 
    & 0.9 & 0.55 & 2.32$\pm$0.63 \\
    
    Infeasible image goal 
    & \xmark 
    & 1 & 0.95 & 15.32$\pm$4.53 
    & - & - & - 
    & - & - & - \\
    
    \bottomrule
\end{tabular}%
}
\label{tab:system_eval}
\end{table*}

\begin{figure*}[!t]
  \includegraphics[width=\linewidth]{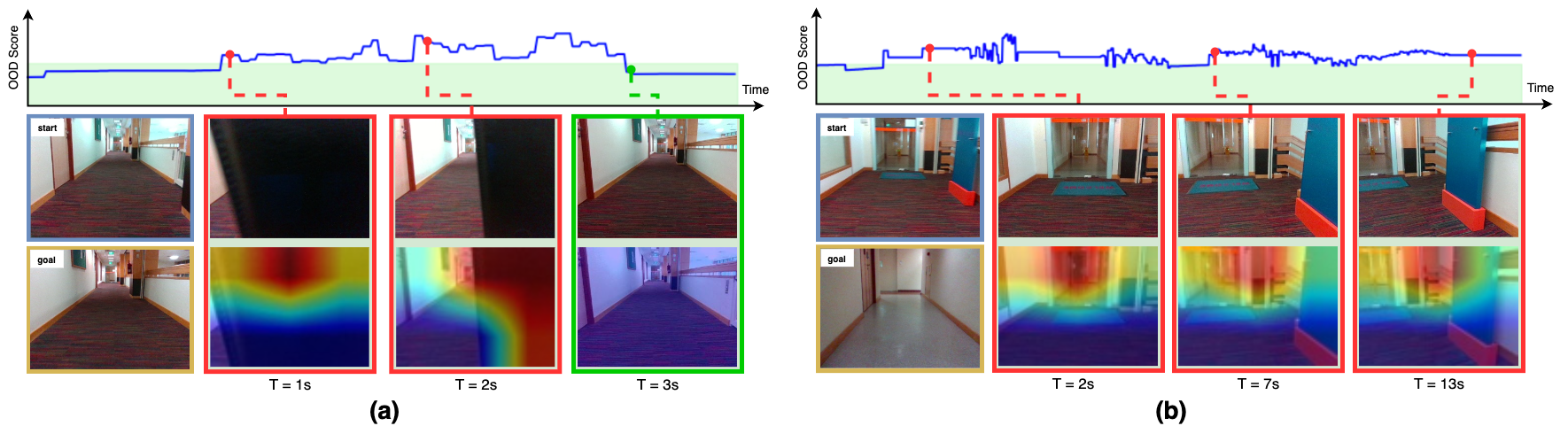}
  \centering
  \caption{\textbf{Informed failure recovery with \frgnm{}.} (\textit{a}): Robot is blocked by a pedestrian passing close by in front. It first backtracks (T=1s) to avoid the pedestrian, recognises the pedestrian on its right (T=2s) and pivots in the opposite direction to recover (T=3s). (\textit{b}): Robot is issued an infeasible goal behind the glass door. It perturbs locally to various directions, but finds itself always blocked by obstacles, thus terminating and seeking help.}
  \label{fig:gnm_qualitative}
\end{figure*}

\begin{figure*}[!t]
  \includegraphics[width=\linewidth]{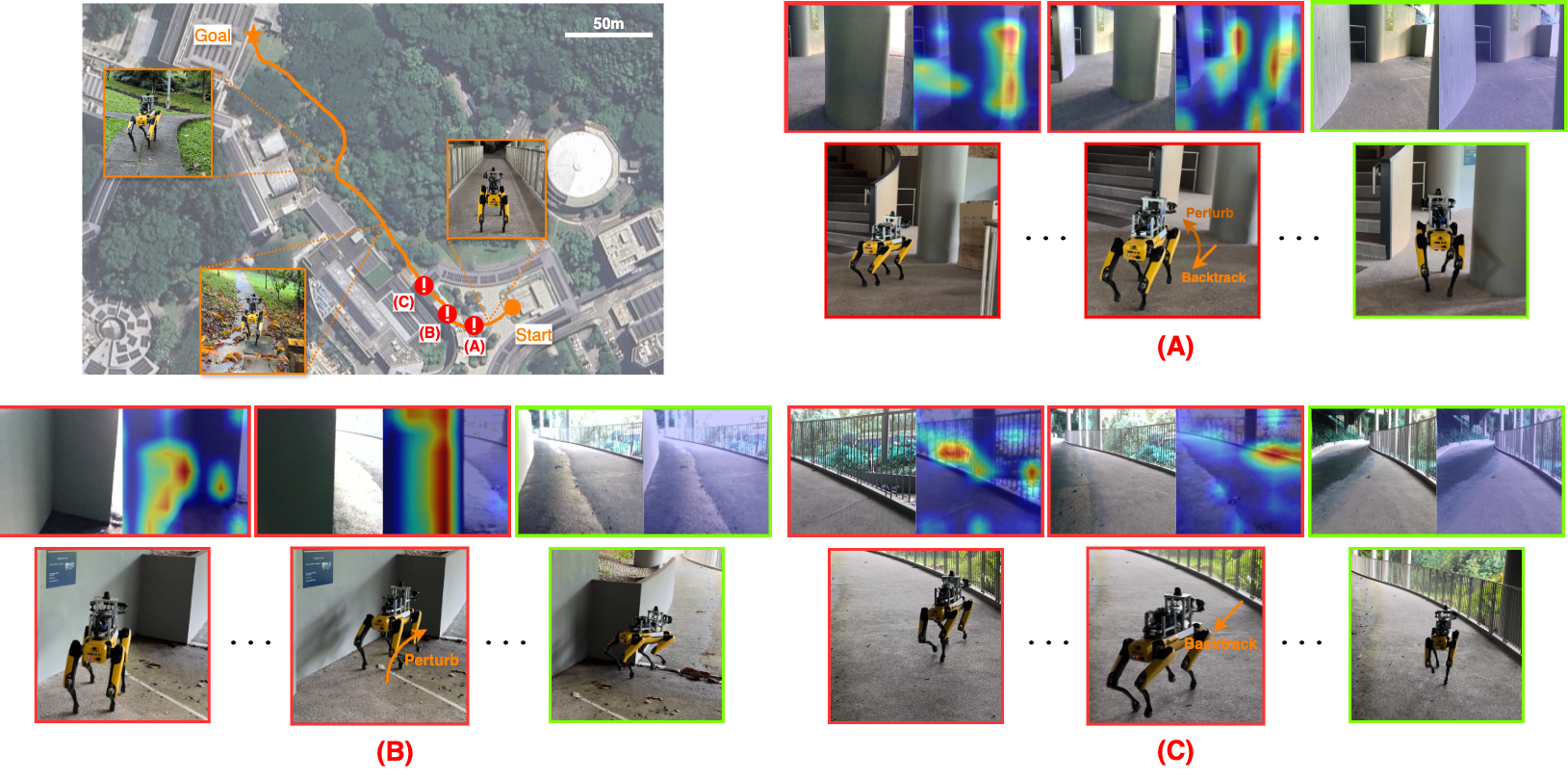}
  \centering
  \caption{\textbf{Informed recovery enables long-range navigation with \frdec{}.} The robot navigates a 300m route starting indoors then crossing a park. It recovers from the three failures shown: \textbf{(A)}: stuck in local minima due to clutter, has to backtrack and perturb to find a path; \textbf{(B)}: stuck in a corner, has to perturb to the right to continue on the path; \textbf{(C)}: heading towards a railing, unseen during training, has to backtrack and align itself along the path.}
  \label{fig:long_range_results}
  \vspace{-1.0em}
\end{figure*}
\textbf{\fr{} effectively handles diverse real-world failures.}
\autoref{tab:system_eval} shows that \fr{} enables high detection and handling success across varied failure types, supporting \textbf{H2}. It is effective at recovering from failures or correctly requesting takeover for irrecoverable failures. It performs especially well in irrecoverable scenarios, where consistent, robust \ood{} detection while the robot perturbs is needed to appropriately terminate recovery efforts and request human help. While overall detection is stable, \frdec{} occasionally fails in local minima due to spurious heatmap activations that guide the robot back into the minima. The exploration strategy can be improved to mitigate this by adding memory or temporally filtering the heatmap. \fr{} also responds effectively to close-range dynamic obstacles, selecting timely corrective actions. Though \frgnm{} shows slower recovery times likely due to gradual KL score decay, it still quickly responds with suitable behaviours like backtracking or perturbing away from obstacles (\autoref{fig:gnm_qualitative}).


\textbf{\fr{} works across policy architectures.} 
\autoref{tab:system_eval} highlights \fr{}'s consistent and strong performance across two policies (GNM and DECISION) despite their differing architectures and failure modes, supporting \textbf{H3}. \frgnm{}’s effectiveness on infeasible image goals (\ref{failure:goal}) shows that the VIB mechanism captures architecture-specific issues. Both \frgnm{} and \frdec{} recover from common failures similarly well, underscoring \fr{}’s generalisability. A factor likely influencing performance is the resolution of the visual encoder’s feature map. \frdec{} (\autoref{fig:long_range_results}) has higher-resolution feature maps than GNM (\autoref{fig:gnm_qualitative}), enabling finer recognition, possibly causing GNM's poorer performance with close-range dynamic obstacles (\ref{failure:dyn}).


\textbf{\ood{} recognition enables efficient, effective recovery.}
Directional recognition information from the heatmap $\locelem{}_t$ improves recovery performance compared to uninformed recovery that randomly selects corrective actions (\undec{}).
\frdec{} consistently outperforms \undec{} over all failure types, especially recovery from close-range dynamic obstacles (\ref{failure:dyn}) where timely and informed actions are critical. While \undec{} can occasionally recover in static environments via random perturbation, dynamic scenarios demand precise recognition. Thus, \frdec{} succeeds in 17/18 detected cases vs. 11/18 for \undec{}. 










\section{Conclusion}

We argue that \textbf{informed failure resilience} in IL policies extends beyond detection to encompass recognising failure sources and autonomously recovering from them. This paper introduces \textbf{\fr{}}, a general framework for building failure-resilient visual navigation policies \textit{without any failure or recovery demonstrations}. \fr{} enables \ood{} detection and recognition grounded in the policy's latent space, and pairs the policy with a separate recovery policy informed by real-time \ood{} recognition results. It does not change the policy’s original
encoder or decoder, adding only two lightweight linear layers
to parameterise the stochastic latent representation. \fr{} generalises across two distinct policy architectures, and a range of diverse failure modes across different scenes, enabling an IL policy to succeed on a challenging long-range navigation case study without intervention. This shows that \fr{} is a clear step toward safe, robust navigation with IL. We will extend it to policies with more advanced visual encoders~\cite{leem2024attention}, and to more sophisticated recovery strategies.

\section*{Acknowledgments}
This work is supported by the NUS President's Graduate Fellowship and DSO Graduate Fellowship.







\bibliographystyle{IEEEtran}
\bibliography{references}


\appendix
\renewcommand{\thesubsubsection}{\thesubsection.\arabic{subsubsection}}
\renewcommand{\thesubsubsectiondis}{\thesubsubsection}

\subsection{Coverage in conformal prediction}
\label{app:coverage}
Our described procedure for functional CP satisfies necessary exchangeability assumptions for coverage. Diquigiovanni et al.~\cite{diquigiovanni2025functionalcp} show that coverage depends on exchangeability of the nonconformity scores, in this case the KL scores. Since our procedure chunks trajectories into non-overlapping segments, with KL scores computed solely from observations within each segment, without any dependence on other segments, exchangeability of the KL scores is satisfied.

\subsection{\ood{} recognition with Grad-CAM++}
\label{app:gradcampp}

While the original \fr{} framework used vanilla Grad-CAM for recognition, many variants of Grad-CAM have been developed. We extend \fr{} to use Grad-CAM++~\cite{chattopadhyay2018gradcampp}, which generalises Grad-CAM to apply location-specific weights to the gradients, improving localisation performance and highlighting salient regions more clearly. 

A limitation of Grad-CAM~\cite{selvaraju2019gradcam} is that it weights each feature map by the spatial average of the gradients of the target class score with respect to that map, which can dilute the contribution of a feature that responds strongly at only a few locations, such as a small instance of the
target class. As shown in \autoref{fig:splash} and \autoref{fig:det_loc_qualitative}, \ood{} features are often localised regions embedded in in-distribution scenes, making Grad-CAM a key bottleneck in improving recognition performance. Grad-CAM++~\cite{chattopadhyay2018gradcampp} instead computes each map's weight as a sum of positive gradients with location-specific coefficients. The resulting heatmap can cover more of an object and highlight multiple instances of the same class, thereby improving recognition performance.

Specifically, Grad-CAM++ gives each location its
own coefficient. Let $D^n_{ij,k}=\partial^n Y^c/\partial (A^k_{ij})^n$ denote the $n$th derivative with respect to a single activation. The coefficient at $(i,j)$ is
\begin{equation}
    \alpha_{ij}^{kc} =
    \frac{D^2_{ij,k}}
         {2D^2_{ij,k} + D^3_{ij,k}\sum_{a,b} A^k_{ab}}.
\end{equation}

These coefficients weight the positive first-order gradients to give the channel weight $w_k^c$.
\begin{equation}
    w_k^c = \sum_{i,j}\alpha_{ij}^{kc}\max(0,D^1_{ij,k})
\end{equation}
The heatmap $L^c$ then combines the feature maps using those weights.
\begin{equation}
    L^c_{ij} = \max\!\left(0,\sum_k w_k^c A^k_{ij}\right).
\end{equation}
By allowing gradient contributions to vary across locations,
Grad-CAM++ can highlight multiple instances of a class and cover
more of an object's extent.

\begin{figure*}[!t]
\centering
  \includegraphics[width=\linewidth]
  {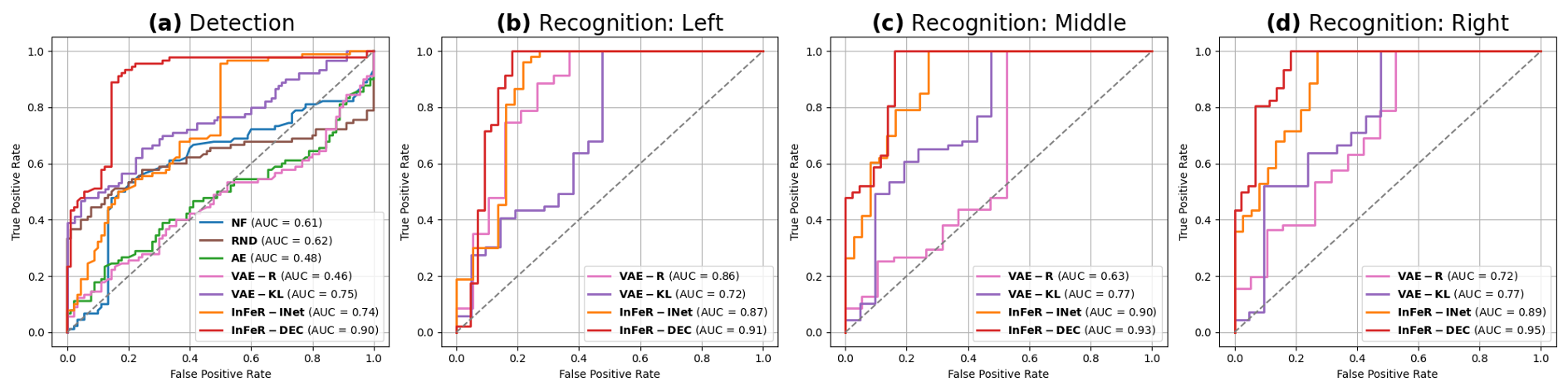}
  \caption{\textbf{Detection and recognition results for improved \frdec{}.} We improve \fr{} to use Grad-CAM++~\cite{chattopadhyay2018gradcampp} for \ood{} recognition instead of the original Grad-CAM used in \frdec{}. Grad-CAM++ enables noticeably better performance for recognition, enabling both \frinet{} and \frdec{} to outperform baselines on this task \textbf{(b)}-\textbf{(d)}.} 
  \label{fig:gradcampp}
\end{figure*}

\autoref{fig:gradcampp} \textbf{(b)}-\textbf{(d)} shows that Grad-CAM++ significantly improves recognition performance with \fr{}, in particular increasing AUC scores for \frinet{} by a large margin. Qualitatively, Grad-CAM++ yields more pronounced and clearer recognition results, while being easy to integrate into the existing \fr{} system. We note that many other improved variants of Grad-CAM have been developed beyond Grad-CAM++, including variants optimised for specific architectures like vision transformers~\cite{leem2024attention}, most of which can serve as direct replacements for Grad-CAM in \fr{}, enabling effective recognition with \fr{} across a wide range of policies.

\end{document}